\documentclass[10pt,twocolumn,letterpaper]{article}

\usepackage{wacv}              

\usepackage{graphicx}
\usepackage{amsmath}
\usepackage{amssymb}
\usepackage{hyperref}
\hypersetup{colorlinks = true}
\usepackage{booktabs}
\usepackage{makecell}

\newcommand\extrafootertext[1]{%
    \noindent
    \bgroup
    \renewcommand\thefootnote{\fnsymbol{footnote}}%
    \renewcommand\thempfootnote{\fnsymbol{mpfootnote}}%
    \footnotetext[0]{#1}%
    \egroup
}

\usepackage[capitalize]{cleveref}
\crefname{section}{Sec.}{Secs.}
\Crefname{section}{Section}{Sections}
\Crefname{table}{Table}{Tables}
\crefname{table}{Tab.}{Tabs.}

\newcommand{\mocap}{\mbox{MoCap}\xspace}
\newcommand{\humor}{\mbox{HuMoR}\xspace}
\newcommand{\nemf}{\mbox{NeMF}\xspace}
\newcommand{\sothree}{SO(3)}

\newcommand{\methodname}{HMP\xspace}
\newcommand{\pymafx}{PyMAF-X\xspace}
\def\rootAlignedErr{RA-MPJPE\xspace}
\def\procrustesAlignedErr{PA-MPJPE\xspace}
\def\rootAlignedAccErr{RA-ACC\xspace}

\def\wacvPaperID{1849} 
\def\confName{WACV}
\def\confYear{2024} 

\begin{document}

\pagenumbering{arabic}

\title{\methodname: Hand Motion Priors for Pose and Shape Estimation from Video}

\author{Enes Duran\textsuperscript{1, 2} \quad 
        Muhammed Kocabas\textsuperscript{1, 3} \quad
        Vasileios Choutas\textsuperscript{1, \footnotemark[2]{}}  \quad
        Zicong Fan\textsuperscript{1, 3} \quad 
        Michael J. Black\textsuperscript{1} \quad  \\
        \small{
\textsuperscript{1}MPI for Intelligent Systems, T\"ubingen, Germany
\textsuperscript{2}University of T\"ubingen, Germany
\textsuperscript{3}ETH Z\"urich, Switzerland} 
\\ 
\small{\textsuperscript{*}Corresponding author: \href{mailto:enes.duran@tue.mpg.de}{enes.duran@tue.mpg.de}}}
\twocolumn[{%
	\renewcommand\twocolumn[1][]{#1}%
	\maketitle   
	\begin{center}
		\newcommand{\teaserwidth}{\textwidth}
		\vspace{-0.15in}
		\centerline{
			\includegraphics[width=\teaserwidth,clip]{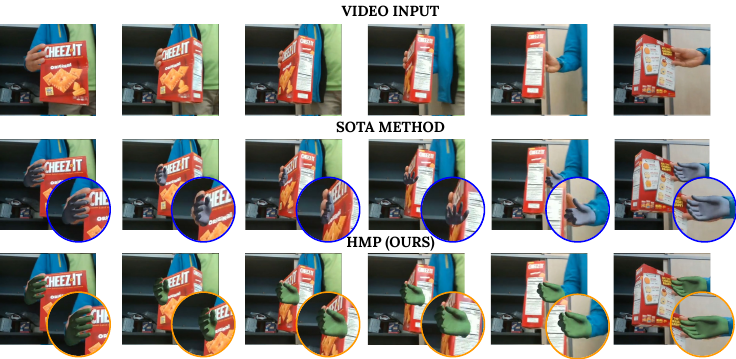}
		}
		\vspace{-2ex}
		\captionof{figure}{Given challenging hand interaction videos (top), a recent state-of-the-art hand pose estimation approach~\cite{pymafx2022} (middle), fails to produce accurate 3D hand poses. To address this, we exploit a large-scale motion-capture dataset AMASS~\cite{AMASS:2019} to train a motion prior and use latent optimization to recover hand pose from videos. Our model \methodname (bottom) is robust to occlusion and produce temporally stable results, outperforming previous work on standard benchmarks.}
  

		\vspace{-0.1in}
		\label{fig:teaser}
	\end{center}%
}]

\maketitle
\begin{abstract}
Understanding how humans interact with the world necessitates accurate 3D hand pose estimation, a task complicated by the hand's high degree of articulation, frequent occlusions, self-occlusions, and rapid motions. While most existing methods rely on single-image inputs, videos have useful cues to address aforementioned issues.  
However, existing video-based 3D hand datasets are insufficient for training feedforward models to generalize to in-the-wild scenarios. On the other hand, we have access to large human motion capture datasets which also include hand motions, \eg AMASS. Therefore, we develop a generative motion prior specific for hands, trained on the AMASS dataset which features diverse and high-quality hand motions. 
This motion prior is then employed for video-based 3D hand motion estimation following a latent optimization approach. 
Our integration of a robust motion prior significantly enhances performance, especially in occluded scenarios. It produces stable, temporally consistent results that surpass conventional single-frame methods. We demonstrate our method's efficacy via qualitative and quantitative evaluations on the HO3D and DexYCB datasets, with special emphasis on an occlusion-focused subset of HO3D. Code is available at 
\href{https://hmp.is.tue.mpg.de}{https://hmp.is.tue.mpg.de}
\end{abstract}
{%
\renewcommand{\thefootnote}{$\dagger$}
\footnotetext{Work done at MPI, now at Google}
}

\section{Introduction}
\label{sec:intro}



Hands often serve as our primary mean for manipulating objects and engaging with our surrounding environments. Therefore, accurately reconstructing the 3D poses and shapes of hands from RGB images plays a crucial role in a range of applications including human--computer interaction, augmented/virtual reality (AR/VR), robotics, biomechanics, and animation. 
Despite years of research in this direction, this task is still challenging due to the high degree of articulation, occlusion caused by hand--object interactions, self-occlusion, and rapid motion inherent to hand movements. 

Existing methodologies predominantly investigate the estimation of hand pose and shape from single images~\cite{zimmermann2017learning,iqbal2018hand, yang2019disentangling,lin2021end,li20143d,park20163d,spurr2020weakly,cai2019exploiting,mueller2018ganerated,cai2018weakly,fan2021digit}. However, such approaches tend to generate temporally-inconsistent reconstructions of hand motion. They are often plagued by jitter, missing predictions, and produce noisy motion results (see middle row of Fig.~\ref{fig:teaser}).

In contrast to the aforementioned scenarios, videos serve as a rich source of data for hand motion analysis. Unlike single images, videos contain temporal information that can help to predict coherent hand reconstructions throughout time by learning correlations between time-adjacent frames. They encapsulate a wealth of cues related to hand motion that could improve hand pose and shape estimation. However, this valuable aspect remains largely under-explored, with very few attempts~\cite{yang2020seqhand,Fu2023DeformerDF,ziani2022tempclr,hasson2020leveraging} being made to leverage video data. Most recently, Fu \etal~\cite{Fu2023DeformerDF} introduced a feedforward model which takes a video as input and reconstructs the observed hand sequence. However, this method has very limited generalization capability since existing video-based 3D hand motion datasets~\cite{ho3d,chao2021dexycb} are limited in terms of the number of subjects and background diversity. On the other hand, large motion capture datasets \eg AMASS~\cite{AMASS:2019} contain accurate and diverse 3D hand pose annotation, but they do not contain images. 
 
Motivated by these observations, our key insight is that we can leverage existing \mocap datasets to build a robust, generative 3D hand motion prior and use this generative motion prior for 3D hand pose and shape estimation from monocular videos. After training on the large AMASS motion capture dataset, we use \methodname(Hand Motion Priors) as a motion prior at test time for 3D hand pose and shape estimation from noisy and partial observations \eg RGB videos and 2D or 3D joint sequences. In particular, we introduce a latent optimization framework which interacts with \methodname to estimate the parameters of hand motion. This
interaction happens by parameterizing the motion in the latent space of \methodname, and by using \methodname priors in order to regularize the optimization towards the space of plausible motions.

Experimental results for 3D hand pose and shape estimation on two existing hand pose estimation video datasets, DexYCB~\cite{chao2021dexycb} and HO3D~\cite{ho3d}, show that our method outperforms state-of-the-art methods. To analyze our method's robustness to occlusion, we curate an occlusion-heavy test set from HO3D which we name HO3D-OCC and demonstrate that our approach is more robust to occlusions than existing approaches. Further, we show that our method surpasses traditional motion priors \eg Gaussian Mixture Models, PCA-based temporal priors, as well as a direct optimization hand pose and shape parameters. 

In summary, our contributions include: 

\begin{itemize}
    \item We introduce a generative hand motion prior learned from a large-scale \mocap dataset AMASS~\cite{AMASS:2019}.
    \item We present a latent optimization-based method for accurate hand pose and shape estimation from monocular videos.   
    \item We demonstrate that our method reconstructs more accurate 3D hand motion under partial or heavy occlusions thanks to our robust generative motion prior. We highlight this on HO3D-OCC, an occlusion-specific subset of HO3D dataset.
    \item We show that our framework allows us to perform better hand reconstruction results compared to traditional temporal priors or direct optimization of hand pose and shape.
\end{itemize}

\section{Related Work}
\label{sec:formatting}

\subsection{Hand Pose Estimation From a Single Image }
Methods estimating 3D hand pose from single images can be split into \textit{model-free} and \textit{model-based} approaches.

Model-free methods~\cite{zimmermann2017learning, yang2019disentangling, lin2021end, li20143d, park20163d, spurr2020weakly} directly estimate the hand pose by predicting 3D joint positions \cite{cai2019exploiting, zimmermann2017learning, doosti2020hope, spurr2018cross, tekin2019h+} or joint heatmaps \cite{iqbal2018hand, mueller2018ganerated, cai2018weakly, fan2021digit}. For instance, Zimmermann \etal \cite{zimmermann2017learning} propose the first convolutional network to detect 2D hand joints and lift them into the 3D space with an articulation prior. Iqbal \etal \cite{iqbal2018hand} introduce a 2.5D representation allowing to make use of supplementary depth supervision. Likewise, Spurr \etal \cite{spurr2020weakly} present biomechanical constraints to refine the pose predictions on 2D supervised data. These methods require abundance of annotated data to train due to the lack of 3D priors. 

MANO~\cite{MANO:SIGGRAPHASIA:2017} is a parametric model of hands. In MANO, the hand is parameterized by pose and shape parameters. The pose parameters define the articulation of the hand, including finger bending and other movements, while the shape parameters define the overall structure and morphology of the hand. 
Several model-based approaches \cite{jointho, semihandobj, baek2019pushing, boukhayma20193d, zhang2019end, fan2021digit, fan2023arctic} directly predict MANO parameters. The 3D hand joint and mesh vertex coordinates are computed from the MANO parameters using linear blend skinning. Zhang \etal \cite{zhang2019end} introduce a framework that harnesses a differentiable re-projection loss for accurate hand mesh recovery. Similarly, Hasson \etal \cite{jointho} use a contact loss that ensures the interaction between the hand and any object appears realistic in predictions. Despite the notable advances achieved by image-based techniques, their results are still not temporally consistent due to occlusions and motion blur present in single frames.


\subsection{Temporal Hand Pose Estimation }

Recent methods \cite{semihandobj, yang2020seqhand, spurr2021adversarial, hasson2020leveraging, handocc, ziani2022tempclr} attempt to leverage temporal data from videos to improve the hand pose estimation performance. %
Hasson \etal~\cite{hasson2020leveraging} use the photometric consistency between the re-projected 3D hand predictions and the optical flow of adjacent frames as supervision. %
Liu \etal~\cite{semihandobj} train an initial model on an annotated dataset, and deploy it on a large-scale video dataset to collect pseudo-labels. They use pseudo-labels to train a single frame model. %
Meanwhile, Ziani \etal~\cite{ziani2022tempclr} use temporal constrastive learning to learn features that are robust to occlusion and motion blur. They also demonstrate the performance of learned features on a temporal model similar to VIBE~\cite{vibe}, a full-body human pose and shape estimation method. %
Fu \etal employ a transformer-driven architecture to process temporal relationship between the input video frames \cite{Fu2023DeformerDF}, resulting in a temporally coherent and accurate hand pose estimation. A limitation of this method is their reliance on video hand pose datasets for training, but the limited subject and background diversity in current datasets impedes the generalizability of methods which take video as input.

In contrast, our method makes use of AMASS motion capture dataset to learn a robust motion prior and fits the latent code of this motion prior to 2D hand keypoint estimations estimated by off-the-shelf algorithms. This makes it more general and flexible compared to existing works relying on video inputs.

\subsection{Motion Prior Models}

Given the absence of motion prior models specific for hands, we turn our attention to methods focused on modeling human body movements. %
There has been a significant amount of research on 3D human dynamics for various tasks, including motion prediction and synthesis~\cite{fragkiadaki2015recurrent,jain2016structural,li2017auto,martinez2017human,villegas2017learning,pavllo2018quaternet,aksan2019structured,gopalakrishnan2019neural,yan2018mt,barsoum2018hp,yuan2019diverse,yuan2020dlow,yuan2020residual,cao2020long,petrovich2021action,hassan2021stochastic}. Recently, human pose estimation methods have started to incorporate learned human motion priors to help resolve pose ambiguity~\cite{vibe,rempe2021humor,zhang2021learning}. Motion-infilling approaches have also been proposed to generate complete motions from partially observed motions~\cite{hernandez2019human,kaufmann2020convolutional,harvey2020robust,khurana2021detecting}. Diffusion models~\cite{sohl2015deep} have also been used as priors for motion synthesis and infilling~\cite{tevet2022human,zhang2022motiondiffuse,yuan2023physdiff,huang2023diffusion}. Rempe \etal~\cite{rempe2021humor} train an autoregressive VAE-based motion prior on AMASS dataset, called \humor. They use \humor as a motion prior at test time for 3D human perception from noisy and partial observations across different input modalities such as RGB videos and 2D/3D joints. It is computationally expensive to perform latent optimization since \humor is autoregressive. Neural Motion Fields (\nemf) express human motion as a time-conditioned continuous function and demonstrate superior motion synthesis performance \cite{he2022nemf}.
Our approach extends NeMF by leveraging it as a motion prior for hands. 
\begin{figure*}
    \centerline{	\includegraphics[width=\textwidth,clip]{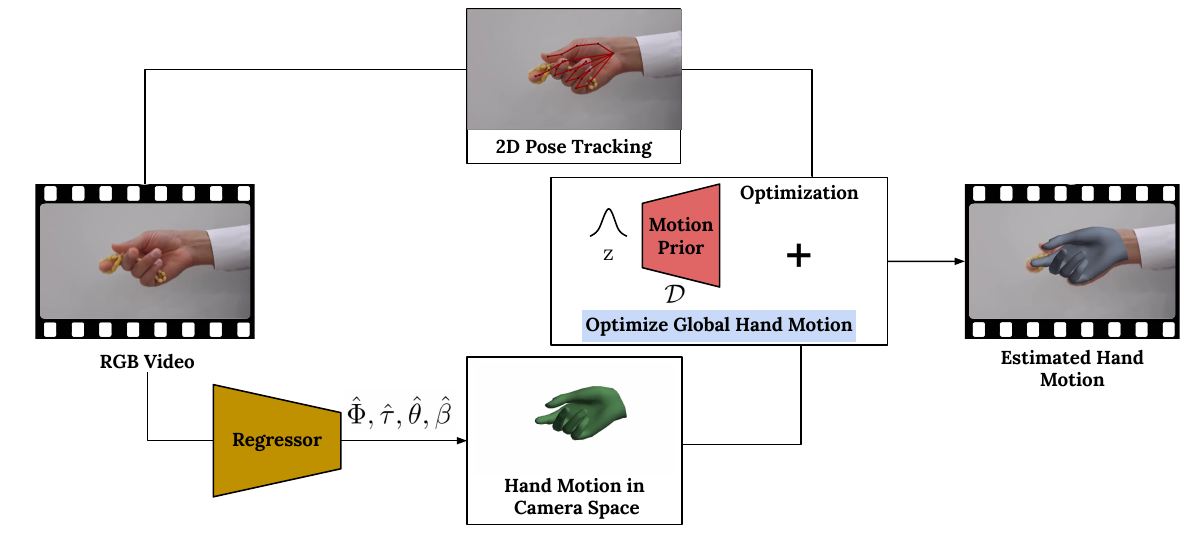}}
    \vspace{-0.1in}
    \caption{\textbf{\methodname method overview:} Given a video of hand, we first use off-the-shelf methods to obtain initial 2D hand keypoints and MANO hand pose and shape parameters via the regressor. We propose a latent optimization framework that optimizes the hand motion to reduce 2D pose errors and increase motion likelihood under hand motion prior. The final output is temporally stable hand motion.}
    \vspace{-0.08in}
    \label{fig:method_overview}
\end{figure*}

\section{Method}

Our method \methodname consists of two phases (\cref{fig:method_overview}):
In the initialization phase, it detects hand bounding boxes, 2D hand keypoints, and initialize MANO hand pose and shape estimates (~\cref{sec:initialization}) from video frames.
In the multi-stage optimization phase (~\cref{sec:latent_optimization}), it then refines those estimates in a video by enforcing hand motion prior constraints. 

\subsection{Preliminaries}
\label{sec:preliminaries}

\methodname takes as input a video of $T$ frames $ I\!=\!\{I_1, ..., I_T \}$. The camera is assumed to be static, \ie $\mathbf{R}_\textrm{cam} = \mathbb{I}$ and $\mathbf{T}_\textrm{cam} = [0, 0, 0]$ where $\mathbf{R}_\textrm{cam} \in \sothree, \mathbf{T}_\textrm{cam} \in \mathbb{R}^{3}$. %
The hand motion in global coordinate system $\mathbf{Q}=\left\{Q_t=\left\{\Phi_t, \tau_t,\theta_t, \beta\right\}\right\}_{t=0}^{T}$ consists of global orientation $\Phi_t \in \sothree$, global translation $\tau_t \in \mathbb{R}^{3}$, hand pose $\theta_t \in \mathbb{R}^{15\times3}$, and hand shape $\beta \in \mathbb{R}^{10}$ for all visible timesteps $t$. 
We use MANO model to represent hand meshes in time \cite{MANO:SIGGRAPHASIA:2017}. Similar to parametric body models, MANO model
outputs a triangulated hand mesh $ V_t \in \mathbb{R}^{778\times3}$ for each timestep $t$ derived from hand motion $\mathbf{Q}$.

Existing 3D hand pose datasets such as HO3D~\cite{ho3d}, DexYCB~\cite{chao2021dexycb} contain images with hands, but they do not have sufficient data with diverse and accurate 3D hand pose annotation.
On other hand, large-scale motion capture datasets such as AMASS~\cite{AMASS:2019} have highly-diverse 3D hand motion data captured in accurate mocap setups, but they do not contain images.
Our key insight is to leverage large-scale motion capture datasets to address the data scarcity problem. 
We first train a hand motion model on AMASS, which learns a prior model  on natural hand motion.
We then introduce a novel optimization-based framework for recovering hand motion by leveraging this motion prior ( \cref{sec:latent_optimization}). 
In this formulation, our method can work with any pose regressors and 2D hand keypoint estimators in plug-and-play fashion (\cref{tab:ho3d_sota,tab:ho3d_ablation}). 
  
\subsection{Initialization}
\label{sec:initialization}
We first obtain bounding boxes for hands using an off-the-shelf hand tracking model~\cite{mmpose2020}.
For each bounding box, we estimate
hand pose and shape in global coordinates 
${\hat{\textbf{Q}}}\!=\!\{\hat{Q}_t\}_{t=0}^{T}$,
where  $\hat{Q}_t = \{\hat{\Phi}_t, \hat{\tau}_t, \hat{\theta}_t,  \hat{\beta}_t\}$,
using \pymafx~\cite{pymafx2022}.
Hand bounding box detection methods often fails when the hand is occluded by objects or during two-hand interactions. To obtain initialization for such frames we perform spherical interpolation (SLERP) for the cases where the bounding box confidence is lower than a threshold.
Our optimization starts from this initial condition. 

For 2D keypoints, we use keypoints combined from MediaPipe and \pymafx \cite{lugaresi2019mediapipe, pymafx2022}. If MediaPipe does not have a detection for a timestep we project 3D joints obtained from \pymafx to the image space and use it as the keypoint source. We experimented with different keypoint estimation methods and empirically found that it is best to blend keypoints from Mediapipe and \pymafx (see Sec \ref{sec:experiments}). This keypoint blending approach aims to combine strengths of each keypoint source for more accurate estimation. We refer to SupMat for more details.

\subsection{Hand Motion Prior}
\label{sec:motion_prior}
Our objective is to build a motion prior to ensure the estimated hand motion's plausibility and to constrain the solution space during motion optimization. To achieve this, we employ a variational autoencoder (VAE)~\cite{kingma2013auto}. The VAE learns a latent representation, denoted as $\mathbf{z}$, of the hand motion and regularizes its latent code distribution to be a normal distribution.
We want the decoder $\mathcal{D}$ of the VAE to be non-autoregressive for faster sampling while not sacrificing accuracy. Such a design choice becomes pivotal, especially when the motion prior is used iteratively during optimization. Autoregressive motion priors, such as \humor~\cite{rempe2021humor}, tend to be unsuitably slow in handling long motion sequences. %
In contrast, a non-autoregressive decoder can be evaluated for the entire sequence in parallel. To this end, we adopt a Neural Motion Field (\nemf)~\cite{he2022nemf} based decoder to represent body motion
as a continuous vector field of hand poses via a NeRF-style MLP~\cite{mildenhall2020nerf}. \Cref{sec:latent_optimization} discusses the application of \nemf for latent optimization.


Building on the approach from~\cite{he2022nemf}, our system solely models local hand motion using the prior. Specifically, $\mathcal{D}$ is an MLP accepting the latent code ${\mathbf{z}_\theta}$ and a time step $t$ to produce the local hand pose $\hat{\theta}_t$ for the respective time step:

\begin{equation}
    \mathcal{D} : (t, \mathbf{z_\theta}) \rightarrow (\hat{\theta}_t), 
\end{equation}

Here, $\mathbf{z}_\theta$ controls the local pose $\theta$ of the hand. Given a specific $\mathbf{z}_\theta$, the entire sequence can be sampled in parallel by simply varying the values of $t$. To incorporate the motion prior during latent optimization, we optimize the latent code $\{\mathbf{z_\theta}\}$ instead of solely optimizing the local hand motion $\{\theta_t\}_{t=0}^{T}$. We initialize the latent code using the pre-trained encoder $\mathcal{E}$ of the VAE; \ie, $z_\theta{=}\mathcal{E}_\theta(\{\theta\}_{t=0}^{T}$). 

As opposed to to the approach in~\cite{he2022nemf}, we omit the global orientation information during motion prior training. This decision is grounded on the observation that hand global orientation is considerably less constrained. Unlike the body's global orientation, which is motion-limited by gravity, hands have the liberty to move freely in the air. Details for the training process are given in SupMat.  


\begin{table}[]
    \centering
    \resizebox{\linewidth}{!}{
    \begin{tabular}{c | c | c | c } 
    \toprule
         Stage & Variables & Loss terms & Description \\
         \midrule 
          1 & $\Phi, \tau, \beta$ & 
          \makecell{$\mathcal{L}_{o}, \mathcal{L}_\textrm{tr}, \mathcal{L}_{\beta},$ 
          \\
          $\mathcal{L}_{os}, \mathcal{L}_{ts}, \mathcal{L}_{\textrm{2D}} $}   & \makecell{global translation \\ +  rotation} 
             \\ 
        \midrule 
          2 & \makecell{ $ \Phi, \tau, \beta, \textbf{z}_{\theta}$ } & \makecell{$\mathcal{L}_{o}, \mathcal{L}_\textrm{tr},
          \mathcal{L}_{\beta}, 
          \mathcal{L}_{os}, $ \\ $
          \mathcal{L}_{\textrm{2D}}, \mathcal{L}_{MP}$}  & \makecell{+ local hand pose}  \\
          \bottomrule
          
    \end{tabular}
    }
    \vspace{-0.3cm}
    \caption{Multi-stage optimization variables and loss terms.}
    \label{tab:opt_stages}
    \vspace{-0.4cm}
\end{table}
\begin{table*}[h]
    \centering
    \resizebox{0.9\linewidth}{!}{

        \begin{tabular}{l|c|c|c}
        \toprule
        \textbf{Stages} & \textbf{Variables} & \textbf{Loss Function} & \textbf{Loss Coefficients}
        \\
        \midrule 
        Stage-1 & $ \Phi, \tau, \beta$ & $\mathcal{L}_{o}, \mathcal{L}_{\textrm{tr}}, \mathcal{L}_{\beta}, \mathcal{L}_{os}, \mathcal{L}_{ts}, \mathcal{L}_\textrm{2D}$ & $
        lr=0.05,
        \lambda_{o}=3,
        \lambda_{\textrm{tr}}=1,
        \lambda_{os}=1,
        \lambda_{ts}=5,
        \lambda_{\beta}=3,
        \lambda_\textrm{2D}=0.05
        $
        \\
        Stage-2 & $\Phi, \tau, \beta, \textbf{z}_{\theta}$
        & $\mathcal{L}_{o}, \mathcal{L}_{\textrm{tr}}, \mathcal{L}_{\beta},
        \mathcal{L}_{os}, 
        \mathcal{L}_{\textrm{2D}}, \mathcal{L}_{\textrm{MP}}$ & $
        lr=0.05, 
        \lambda_{o}=2, 
        \lambda_{\textrm{tr}}=1, 
        \lambda_{os}=1, 
        \lambda_{\beta}=10, \lambda_\textrm{2D}=0.05, \lambda_\textrm{MP}=300 $
        \\
        \bottomrule 
        \end{tabular} 
    }
   \caption{Multi-stage optimization loss coefficients.  
   }
   \vspace{-0.3cm}
    \label{tab:opt_stages_detailed}
\end{table*}

\subsection{Latent Optimization}
\label{sec:latent_optimization}

This section explains the latent optimization process for estimating hand motion. Our goal is to optimize the variables expressing hand motion: global orientation $\Phi$, global translation $\tau$, shape $\beta$, and the motion prior latent code  $z_{\theta}$.  

\paragraph{Objective Function:} The objective function we aim to minimize is defined as:

\begin{equation}
\begin{split}
\mathcal{L} = \lambda_{o} \mathcal{L}_{o} +  \lambda_{\textrm{tr}} \mathcal{L}_{\textrm{tr}} +\lambda_{\beta}  \mathcal{L}_{\beta} + \lambda_{os} \mathcal{L}_{os} + \lambda_{ts} \mathcal{L}_{ts} \\ + \lambda_{\textrm{MP}} 
 \mathcal{L}_{\textrm{MP}} + \lambda_{\textrm{2D}} \mathcal{L}_{\textrm{2D}},
\end{split}
\end{equation}
where we use seven different objective terms with their corresponding coefficients. Values for these coefficients are given in \cref{tab:opt_stages_detailed}. The first term ensures that the optimized global orientation is close to the initial global orientation through $\mathcal{L}_{\textrm{o}}$:
\begin{equation}
\mathcal{L}_{\textrm{o}}= \sum_{t=0}^{T}  g(\Phi^{}_{t}, \hat{\Phi}^{}_{t})^2.
\end{equation}
where $g$ is the geodesic distance between two rotations. The second term $\mathcal{L}_{\textrm{tr}}$ encourages the global translation not to diverge from its initial values:
\begin{equation}
\mathcal{L}_{\textrm{tr}}= \sum_{t=0}^{T}  \| \tau_{t} -\hat{\tau}_{t} \|_{2}^{2}.
\end{equation}
The third term motivates shape parameters to be close to zero vector through $\mathcal{L}_{\beta}$:
\begin{equation}
\mathcal{L}_{\beta} = \| \beta \|_{2}^{2}.
\end{equation}
The fourth and fifth terms encourage smoothness of the global translation and orientation:
\begin{align}
    \mathcal{L}_{\textrm{os}} &= \sum_{t=0}^{T-1}  g(\Phi_{t+1}, \Phi_t)^2,
    \\
    \mathcal{L}_{\textrm{ts}} &= \sum_{t=0}^{T-1}  \left\| \tau_{t+1} - \tau_{t} \right\|^{2}_{2}. 
\end{align}
The 2D keypoint error term, $\mathcal{L}_{\textrm{2D}}$, constrains our motion to be aligned with the 2D keypoints predicted from detectors:
\begin{equation}
\mathcal{L}_{2 \mathrm{D}}=\sum_{i=1}^{21} \sum_{t \in T_\textrm{detect}} \alpha^{i}_{t} \rho\left(\Pi\left(R_\textrm{cam} J_t^i+ T_\textrm{cam} \right) -\mathrm{x}_t^i \right).
\end{equation}
Here $T_\textrm{detect}$ represents the time-steps where we have the keypoint detection. $J^{i}_{t}$ stands for location of joint $i$ in timestep $t$. $\mathrm{x}_t^i$ is the corresponding detected keypoint in image space, $\Pi$ represents perspective projection to image space using camera intrinsics K, $\alpha^{i}_{t}$ represents the detection confidence for the joint $i$, and $\rho$ is the Geman-McClure function \cite{geman_statistical_1982}. The last term constrains hand motion to be valid by minimizing the negative log-likelihood of the latent code.
\begin{equation}
\mathcal{L}_{\mathrm{MP}}= - \log \mathcal{N}\left(\mathbf{z}_\theta ; \mu_\theta\left(\{\theta_t\}\right), \sigma_\theta\left(\left\{\theta_t\right\}\right)\right) .
\end{equation}
\noindent\textbf{Multi-Stage Optimization:} The estimation of 3D pose and shape from 2D video presents an inherently ill-posed problem. Attempting to optimize all parameters simultaneously can lead to local minima. To mitigate this, we adopt a gradual optimization process that progresses from coarse to fine-grained level. This approach serves the purpose of constraining the optimization problem at each stage.

Our optimization strategy unfolds in two distinct stages. During the initial stage, our objective is to align the initial hand estimations from \pymafx with the video data by optimizing the global orientation $\Phi$, global translation $\tau$, and shape $\beta$ only. In the second stage, we include the motion prior latent code $z$ to optimize the local pose, a phase that involves a more refined level of optimization. We use the Adam optimizer \cite{Adam}.

\noindent\textbf{Occlusion Handling}: For occluded frames, the off-the-shelf methods used for initialization, \ie 2D keypoint and bounding-box detectors, \pymafx, do not provide any results. To robustly handle such frames without detection, we mask the objective terms in corresponding time steps, leaving us to solely optimize the latent code $\mathbf{z}_\theta$ with the observed time steps. Motion prior in such cases behave as an motion infilling method which infer the occluded frames with the cues from visible frames. This is a key part of our method which makes it robust to occlusions.
   
\noindent\textbf{Parallel Optimization:} A natural candidate for motion prior formulation is HuMoR \cite{rempe2021humor}. However its autoregressive formulation renders it impractical for using long sequences. Instead, we aim to have a motion prior suitable for batch optimization. Our architecture is based on a recent work NeMF \cite{he2022nemf}. Its formulation allows parallel optimization, making it applicable to long motion sequences.
\section{Experiments} 
\label{sec:experiments}

\subsection{Datasets and Metrics}

\noindent\textbf{HO3D} is a dataset focused on capturing temporal interactions between hands and objects. This dataset comprises interactions of 10 subjects with 10 distinct YCB objects, all captured from various viewpoints \cite{xiang2017posecnn, ho3d}. The manipulation of handheld objects within this dataset often results in substantial occlusions, posing challenges for analysis. We worked on version-3 of the dataset for evaluation.

\noindent\textbf{HO3D-OCC}: We choose occlusion-specific sequences from HO3D to highlight the performance of different methods under occlusion. This subset is derived from the training segment of HO3D and is comprised of 1736 frames.

\noindent\textbf{DexYCB \cite{chao2021dexycb}:} This dataset contains 10 subjects grasping 20 different objects from YCB-Video dataset \cite{xiang2017posecnn}. Ground-truth values are obtained through an optimization process using hand-annotated 2D keypoints, and multiview RGB-D captures. The sequences are shorter (2-3 seconds) and motions have less articulation in comparison to HO3D~\cite{ho3d}. The default split (\textbf{S0}) is used for evaluation.

\noindent\textbf{AMASS~\cite{AMASS:2019}} is a large dataset of 3D human motion capture curated from various marker-based datasets. Among many datasets included in the AMASS repository, GRAB \cite{GRAB:2020}, TCDHands, and SAMP \cite{hassan2021stochastic} feature hand articulations. We use these datasets to train our motion prior.

\noindent\textbf{Metrics:} We report Procrustes aligned  (\procrustesAlignedErr), root aligned (\rootAlignedErr) Mean-Per-Joint Projection Error in millimeters ($mm$). We also report root aligned acceleration error (\rootAlignedAccErr) in $mm/s^2$. Acceleration error demonstrate the smoothness of estimated motion.


\begin{table}
    \centering
    \resizebox{\linewidth}{!}{
        \begin{tabular}{l|r|r|r}
        \toprule
        & \multicolumn{3}{c}{\textbf{HO3D-v3}} \\
        \textbf{Methods} & \textbf{\procrustesAlignedErr} $\downarrow$ & \textbf{\rootAlignedErr} $\downarrow$ & \textbf{\rootAlignedAccErr} $\downarrow$  
        \\ 
        \midrule 
        TempCLR$^\dagger$ \cite{ziani2022tempclr}  & 10.6 & - & 3.7 
        \\
        HandOccNet$^\dagger$ \cite{handocc} & 9.1 & 24.9 & -  
        \\
        \midrule 
        
        METRO \cite{lin2021end} & 12.1 & 38.7 & 17.4
        \\
        PyMAF-X \cite{pymafx2022} & 10.8  & 29.6 & 9.3
        \\
        \cite{lin2021end} + \methodname (Ours) & 10.8  & 31.3 & 2.4
        \\ 
        \cite{pymafx2022} + \methodname (Ours) & \textbf{10.1} & \textbf{26.7} & \textbf{2.2}
        \\ 
        \bottomrule
        \end{tabular}
        }
   \vspace{-2mm}
   \caption{State-of-the-art comparison on the HO3D-v3 dataset \cite{hampali2021ho3dv3}. Methods denoted with $\dagger$ uses HO-3D as their training dataset.}
   \vspace{-2mm}
   \label{tab:ho3d_sota}
\end{table}

\begin{table}
    \centering
    \resizebox{\linewidth}{!}{
        \begin{tabular}{l|r|r|r}
        \toprule
            & \multicolumn{3}{c}{\textbf{DexYCB}} \\
        \textbf{Methods} & \textbf{\procrustesAlignedErr} $\downarrow$ & \textbf{\rootAlignedErr} $\downarrow$ & \textbf{\rootAlignedAccErr} $\downarrow$  \\ 
        \midrule
        ArtiBoost$^\dagger$ \cite{yang2021ArtiBoost} & - & 12.8 & -
        \\
        Deformer$^\dagger$  \cite{Fu2023DeformerDF} & 5.2 & - & - \\
        \midrule
        PyMAF-X \cite{pymafx2022} & 11.6  &  38.1 & 17.1
        \\
        \cite{pymafx2022} + \methodname (Ours) & \textbf{8.9} & \textbf{34.1} & \textbf{3.6} \\
        \bottomrule
        \end{tabular}
    }
   \vspace{-2mm}
   \caption{State-of-the-art comparison on the DexYCB dataset \cite{chao2021dexycb}. Methods denoted with $\dagger$ uses DexYCB as their training dataset.}
   \vspace{-4mm}
   \label{tab:dexycb_sota}
\end{table}

\subsection{Comparison With the State-of-the-Art}

Our aim is to have a method that generalize well to video from different sources. One way of ensuring that is to use a method that performs best in in-the-wild settings. \pymafx~\cite{pymafx2022} is the current SOTA hand pose and shape estimation method. We use \pymafx as our main baseline and report other results based on it.

Our main goal in this paper is to recover coherent motion of hands. Therefore, we would like to emphasize metrics which measures the quality of the estimated motion \eg \rootAlignedAccErr. Unfortunately such metrics are not available for the evaluation server of the HO3D dataset. Therefore, in \cref{tab:ho3d_sota} we use the training set to report metrics and compare with methods doing so. All results listed on DexYCB \cref{tab:dexycb_sota} use \textbf{S0} subset.

Recent methods, such as Deformer~\cite{Fu2023DeformerDF} and ArtiBoost~\cite{yang2021ArtiBoost}, use HO3D and DexYCB as their primary training datasets. However, given the limited background and subject diversity inherent to HO3D and DexYCB, methods solely trained on these datasets struggle to generalize effectively to in-the-wild videos. In contrast, neither \pymafx nor our motion prior relies on these datasets for training, thereby enhancing their generalization to in-the-wild scenarios. Consequently, directly comparing our method with those trained on HO3D and DexYCB can be challenging. To signify this distinction, we have marked such methods with a $\dagger$ in the corresponding tables. Overall, our method outperforms the existing state-of-the-art (SOTA) techniques on the HO3D and DexYCB datasets. Furthermore, our approach enhances the performance of the \pymafx method, which we employ for initialization, across both datasets.

\begin{figure}[t]
    \centering
    \includegraphics[width=\linewidth]{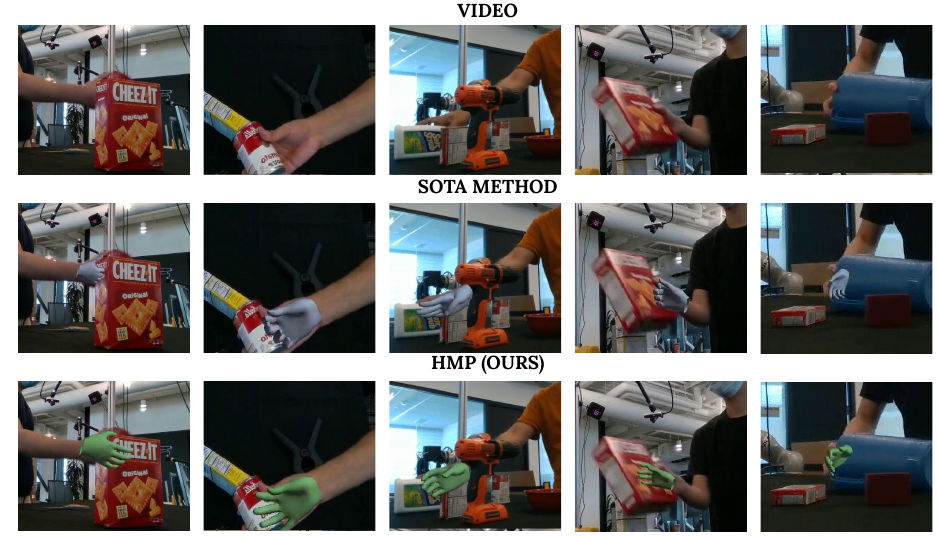}
    \vspace{-7mm}
    \caption{\textbf{3D hand pose and shape estimation on DexYCB videos:} input video (top), \pymafx (middle), \methodname (bottom)} 
    \label{fig:dexycb_qual}
    \vspace{-8mm}
\end{figure}

\begin{figure}[t]
    \centering
    \includegraphics[width=\linewidth]{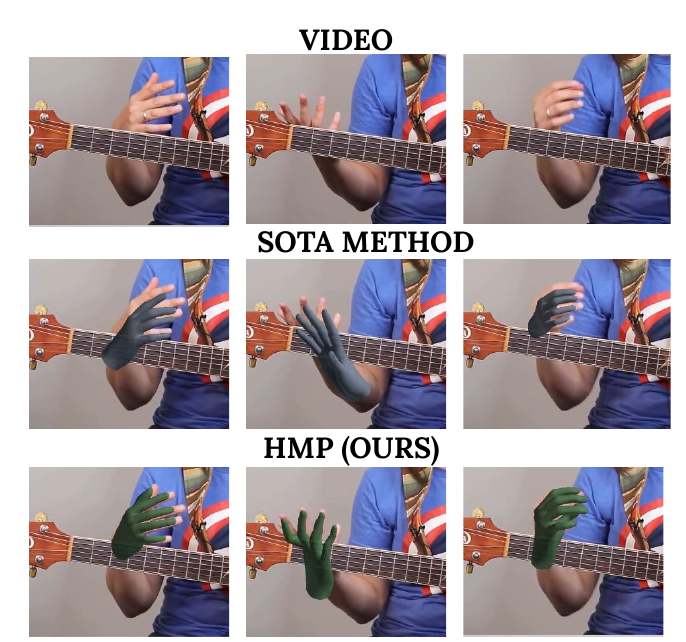}
    \vspace{-7mm}
    \caption{\textbf{3D hand pose and shape estimation on an in-the-wild video:} input video (top), \pymafx (middle), \methodname (bottom)} 
    \label{fig:in_the_wild_qual2}
    \vspace{-6mm}
\end{figure}

\begin{figure*}[t]
    \centering
    \includegraphics[width= 1 \linewidth]{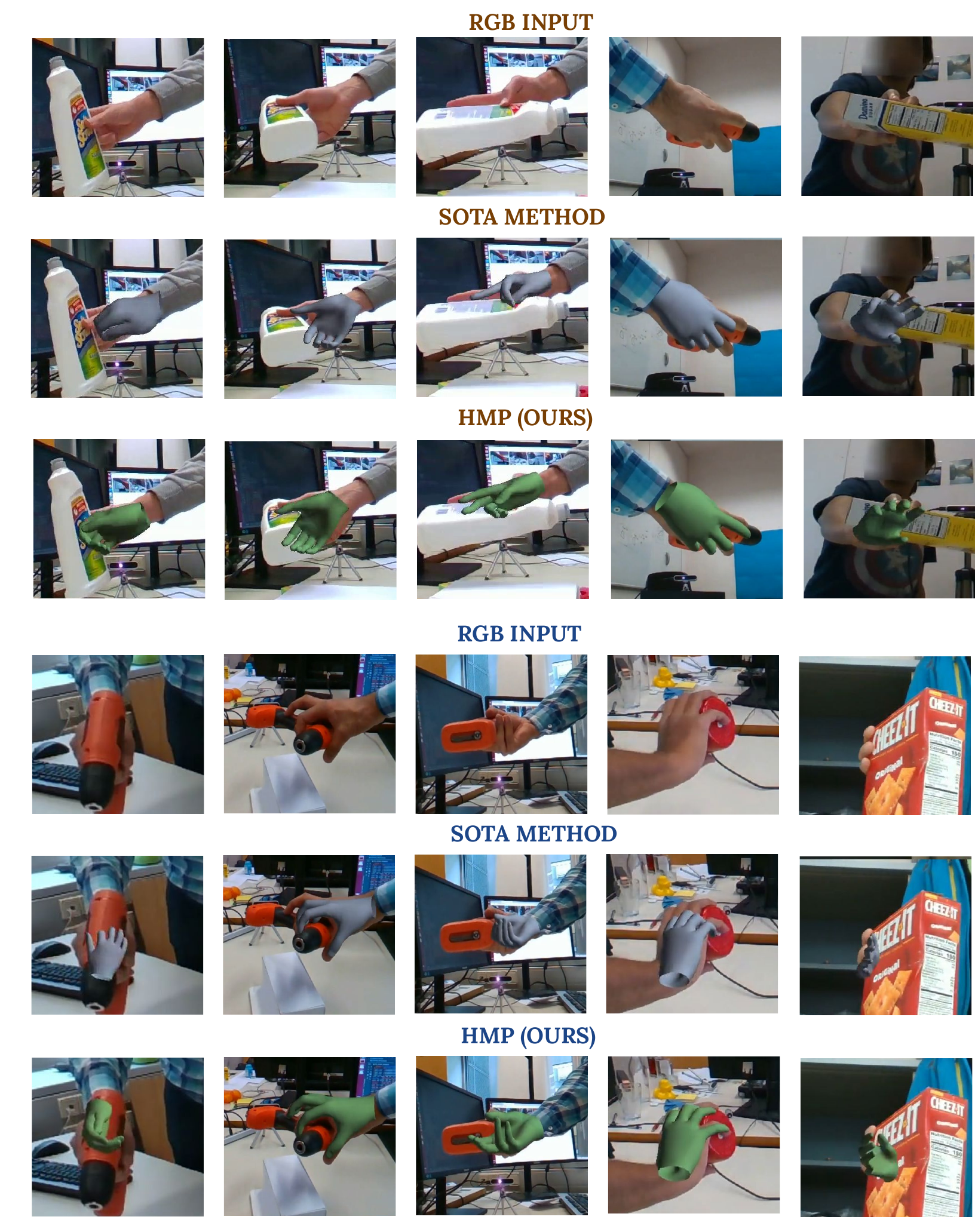}
    \vspace{-10mm}
    \caption{\textbf{3D hand pose and shape estimation on HO3D videos:} input video (top), \pymafx (middle), \methodname (bottom)} 
    \label{fig:ho3d_qual}
\end{figure*}

Additionally, we provide qualitative results DexYCB in \cref{fig:dexycb_qual}, on an in-the-wild video in \cref{fig:in_the_wild_qual2}, and on HO3D in \cref{fig:ho3d_qual}. Please see the SupMat for more results. Compared to \pymafx method, our method is robust to occlusion caused by hand-object interaction. 

To show the applicability of aforementioned plug-and-play fashion in \cref{sec:initialization},
we also report quantitative numbers for optimization with different initialization methods in \cref{tab:ho3d_sota}. 

\begin{table}
    \centering
    \resizebox{\linewidth}{!}
    {
        \begin{tabular}{l|r|r|r}
        \toprule
        & \multicolumn{3}{c}{\textbf{HO3D-v3}}  \\
        \textbf{Methods} & \textbf{\procrustesAlignedErr} $\downarrow$ & \textbf{\rootAlignedErr} $\downarrow$ &  \textbf{\rootAlignedAccErr} $\downarrow$  \\
        \midrule 
        \makecell{PyMAF-X \cite{pymafx2022} \\ + SLERP } & 
        10.7 & 29.4 & 5.9 
        \\
        \midrule
        No Motion Prior & 10.5 & 28.0 & 1.9
        \\
        PCA-based Prior & 13.8 & 31.1 & 10.7
        \\
        GMM-based Prior & 10.4 & 27.5 & 3.4
        \\
        \midrule
        Stage-1 (PyMAF-X) & 10.5 & 26.8 & 2.0              
        \\
        Stage-1 (MediaPipe) & 10.3 & 27.0  & 1.9 
        \\
        Stage-1 (MMPose) & 10.3 & 27.1 & \textbf{1.8}  
        \\
        Stage-1 (Blend) & 10.2 & 27.7 & 1.9
        \\
        Stage-2 (Blend) & \textbf{10.1} & \textbf{26.7} & 2.2 
        \\
        \midrule
        PyMAF-X \cite{pymafx2022} & 10.8  & 29.6 & 9.3 
        \\
        \cite{pymafx2022} + \methodname (Ours) & \textbf{10.1} & \textbf{26.7} & 2.2 
        \\ 
        \bottomrule
        \end{tabular}
        }
   \vspace{-0.2cm}
   \caption{Ablation studies on the HO3D-v3 dataset \cite{hampali2021ho3dv3}.}
   \vspace{-0.6cm}
   \label{tab:ho3d_ablation}
\end{table}

\begin{table}
    \centering
    \resizebox{\linewidth}{!}{
        \begin{tabular}{l|r|r|r}
        \toprule
        & \multicolumn{3}{c}{\textbf{DexYCB}}  \\
        \textbf{Methods} & \textbf{\procrustesAlignedErr} $\downarrow$ & \textbf{\rootAlignedErr} $\downarrow$ & \textbf{\rootAlignedAccErr} $\downarrow$ \\
        \midrule 
        \makecell{PyMAF-X \cite{pymafx2022} + SLERP} & 
        11.5 & 36.5 & 6.0     
        \\
        \midrule
        No Motion Prior & 10.9 & 36.4 & 3.4
        \\
        PCA-based Prior & 16.9 & 41.6 & 15.3
        \\
        GMM-based Prior & 10.8 & 38.7 & 4.8
        \\
        \midrule
        Stage-1 (PyMAF-X) & 10.8 & 35.1 & \textbf{3.4}     
        \\
        Stage-1 (MediaPipe) & 10.8 & 39.1 & \textbf{3.4}    
        \\
        Stage-1 (MMPose) & 10.8 & 39.4 & \textbf{3.4} 
        \\
        Stage-1 (Blend) & 10.8 & 35.5 & \textbf{3.4} 
        \\
        Stage-2 (Blend) & \textbf{8.9} & \textbf{34.1} & 3.6   
        \\
        \midrule
        PyMAF-X \cite{pymafx2022} & 11.6 & 38.1 & 17.1
        \\
        \cite{pymafx2022} + \methodname (Ours) & \textbf{8.9} & \textbf{34.1} & 3.6   
        \\ 
        \bottomrule
        \end{tabular}
        }
   \vspace{-0.2cm}
   \caption{Ablation studies on the DexYCB dataset \cite{chao2021dexycb}.}
   \vspace{-0cm}
   \label{tab:dexycb_ablation}
\end{table}

\begin{table}[h]
    \centering
    \resizebox{\linewidth}{!}{
    \begin{tabular}{l|r|r|r}
        \toprule
        & \multicolumn{3}{c}{\textbf{HO3D-OCC}}   \\
        \textbf{Methods} & \textbf{\procrustesAlignedErr} $\downarrow$ & \textbf{\rootAlignedErr} $\downarrow$ &  \textbf{\rootAlignedAccErr} $\downarrow$
        \\
        \midrule 
        PyMAF-X \cite{pymafx2022} &
        15.3 & 48.9 & 26.0
        \\
        PyMAF-X \cite{pymafx2022} + SLERP & 
        14.4 & 41.3 & 7.9 \\
        \midrule
        Stage-1 & 13.0 & 38.2 & \textbf{2.8} \\ 
        Stage-2 & \textbf{12.6} & \textbf{38.1} & 3.0 \\ 
        \bottomrule
        \end{tabular}
    }
    \vspace{-3mm}
   \caption{Performances on occlusion specific HO3D-OCC subset}
   \vspace{-6mm}
   \label{tab:ho3d_occlusion}
\end{table}

\subsection{Ablation Study}

We report the ablation experiments on HO3D and DexYCB datasets, in \cref{tab:ho3d_ablation,tab:dexycb_ablation} respectively. In this section, we analyze the critical components of our method.

\noindent\textbf{Motion Prior:} In addition to the NeMF-based motion prior we use, we report the results of using no motion prior, a PCA-based motion prior, and GMM-based motion prior, denoted as \textit{No Motion Prior}, \textit{PCA-based prior}, and \textit{GMM-based prior} respectively. In the \textit{No Motion Prior} experiments, we directly optimize the MANO hand pose instead of the latent code of the motion prior. We introduce a pose smoothness term to the pose optimization process, replacing the motion prior likelihood.
Our motion prior trained on AMASS dataset outperforms these motion prior baselines on both the HO3D and DexYCB datasets. 

\noindent\textbf{Multi Stage:} We run ablation studies to demonstrate the results of different stages of our optimization process. We show that Stage-2 which optimizes local pose through latent optimization help to improve results over the Stage-1.

\noindent\textbf{Keypoint Blending:} We analyzed the performance with different 2D hand keypoint detection algorithms: MMPose~\cite{mmpose2020}, MediaPipe~\cite{lugaresi2019mediapipe}, and \pymafx~\cite{pymafx2022}. We also reported a variant where we blend keypoints from MediaPipe~\cite{lugaresi2019mediapipe} and \pymafx~\cite{pymafx2022}. We find out that blending MediaPipe and \pymafx are better 2D hand keypoint detectors compared to MMPose. Blending MediaPipe with the \pymafx give the best results overall.



{}
{}


{}

\section{Conclusion and Discussion}

\label{conclusion}

In this work we propose \methodname, a latent optimization-based method for 3D hand pose and shape estimation from video. Motivated by the fact that existing video-based 3D hand datasets are insufficient for training feedforward models to generalize to in-the-wild scenarios, we develop a generative motion prior specific for hands, trained on the AMASS dataset and then employ motion this prior for video-based 3D hand motion estimation following a latent optimization approach.

Our integration of a robust motion prior significantly enhances performance, especially in occluded scenarios. It produces stable, temporally consistent results that surpass conventional single-frame methods. Our method's efficacy is demonstrated through both qualitative and quantitative evaluations on the HO3D and DexYCB datasets, with special emphasis on an occlusion-focused subset of HO3D. Our method can be used in plug-and-play fashion with any single-stage pose and shape regressor and improves its performance further. Due to this flexibility, unlike existing video hand pose and shape estimation methods, \methodname works on in-the-wilds videos, too.

\noindent\textbf{Limitations:} A limitation of our approach is its reliance on 2D keypoint estimation quality. Existing 2D keypoint predictors can fail under heavy occlusion or motion blur. 

{\small
\noindent\textbf{Acknowledgements:} We thank Nikos Athanasiou, Peter Kulits, and all Perceiving Systems department members for their valuable feedback and insightful discussions. 

\noindent\textbf{Disclosure:} \url{https://files.is.tue.mpg.de/black/CoI_ICCV_2023.txt}
}

{\small
\bibliographystyle{ieee_fullname}

} 
\end{document}


\title{Supplementary Material for \\
\methodname: Hand Motion Priors for Pose and Shape Estimation From Video
}

\maketitle
\newpage
\newpage

\section{Additional Experiments}

\subsection{HO-3D Test Split Results}
We report quantitative results over the HO3D test split in \cref{tab:ho3d_test_sota}. In addition to mean-per-joint-projection error (\procrustesAlignedErr) and vertex-to-vertex error (\procrustesAlignedV) we provide the F-scores after procrustes alignment: \procrustesAlignedFFive, and  \procrustesAlignedFFifteen. Those values are obtained from the official evaluation server using the test set of HO-3D. Since we do not have the ground-truth labels for the test set, we cannot compute and report \rootAlignedAccErr.

Recent methods utilize HO3D and DexYCB as their primary training datasets. However, given the limited background and subject diversity inherent to HO3D and DexYCB, methods solely trained on these datasets struggle to generalize effectively to in-the-wild videos. In contrast, neither \pymafx nor our motion prior relies on these datasets for training, thereby enhancing their generalization to in-the-wild scenarios. Consequently, directly comparing our method with those trained on HO3D and DexYCB can be challenging. To signify this distinction, we have marked such methods with $\dagger$ in the corresponding tables. Overall, our method outperforms the existing state-of-the-art (SOTA) techniques on the HO3D-test split. Furthermore, our approach enhances the performance of the \pymafx method, which we employ for initialization, across both datasets.

\begin{table}[b]
    \centering
    \resizebox{\linewidth}{!}{
        \begin{tabular}{l|r|r|r|r}
        \toprule
        & \multicolumn{4}{c}{\textbf{HO3D-v3}} \\
        \textbf{Methods} & \textbf{\procrustesAlignedErr} $\downarrow$ & \textbf{\procrustesAlignedV2V} $\downarrow$ & \textbf{\procrustesAlignedFFive} $\uparrow$  & 
        \textbf{\procrustesAlignedFFifteen} $\uparrow$  
        \\ 
        \midrule 
        Hasson \etal$^\dagger$ \cite{hasson2020leveraging} & 11.4 & 11.4 & 42.8 & 93.2
        \\
        Hasson \etal$^\dagger$ \cite{hasson19_obman}  & 11.1 & 11.0 & 46.0 & 93.0
        \\ 
        TempCLR$^\dagger$  
\cite{ziani2022tempclr}  & 10.6 & 10.6 & 48.1 & 93.7  
        \\
        Hampali \etal$^\dagger$ \cite{ho3d} & 10.7 & 10.6 & 50.6 & 94.2  
        \\
        Liu \etal$^\dagger$ \cite{semihandobj} & 10.1 & 9.7 & 53.2 & 95.2 
        \\ 
        Deformer$^\dagger$ \cite{Fu2023DeformerDF} & 9.4 & 9.1 & 54.6 & \textbf{96.3}
        \\
        HandOccNet$^\dagger$ \cite{handocc} & \textbf{9.1} &\textbf{8.8} & \textbf{56.4} & \textbf{96.3}
        \\
        \midrule 
        PyMAF-X \cite{pymafx2022} & 11.2  & 11.0 & 47.6 & 92.8
        \\
        \methodname (Ours) & \textbf{10.2} & \textbf{9.9} & \textbf{51.0} & \textbf{94.6}
        \\ 
        \bottomrule
        \end{tabular}
        }
   \caption{State-of-the-art comparison on the test split of HO3D-v3 dataset \cite{hampali2021ho3dv3}. Methods denoted with $\dagger$ uses HO-3D as their training dataset.}
   \label{tab:ho3d_test_sota}
\end{table}

\subsection{Sample Diversity of Different Motion Priors}
We report the sample diversity metrics for PCA-based motion prior, GMM-based motion prior, and our motion prior in \cref{tab:mp_diversity}. All motion priors are trained on the same sequences from the AMASS dataset. We follow the same evaluation criteria as \cite{he2022nemf}. 

\begin{table}
    \centering
    \begin{tabular}{ c | c } 
    \toprule
         MP Type & Diversity $\uparrow$ \\ 
         \midrule
         PCA-based & 5.83
         \\
         \midrule
         GMM-based & 5.79
         \\ 
         \midrule
        \methodname (Ours) & \textbf{5.86} 
        \\
        \bottomrule
          
    \end{tabular}
    \caption{Diversity metrics for different motion prior types}
    \label{tab:mp_diversity}
\end{table}

\section{Additional Qualitative Results \& Remarks}

\textbf{Obtaining 2D Pose Confidence}: We use keypoint confidences to weight $\mathcal{L}_{\textrm{2D}}$. Unfortunately, MediaPipe does not provide separate confidences for hand keypoints \cite{lugaresi2019mediapipe}. To obtain confidence keypoints, we augment with 11 different views through random rotation and scaling. We perform detection on these views and project the results back. For each joint with index $j$ we compute an std value $\sigma_j$:

\begin{equation}
\sigma_{j}^{2} = \frac{1}{N} \sum_{n=1}^{N} (P_{n} - P_{0})^{2}.
\end{equation}
Here $P_{0}$ and $P_{n}$ denotes the original view and n'th view obtained by random scaling and rotation. To discard anomalies, we clip the std values by an upper threshold $\gamma$:
\begin{equation}
\sigma_{j} = \min{(\sigma_{j}, \gamma)}.
\label{eq:std_clip}
\end{equation}
We then compute confidence value as,
\begin{equation}
\alpha_{j} = 1 - \frac{\sigma_{j}}{\gamma}.
\label{eq:std2conf}
\end{equation}
We set $N$=11 and $\gamma$ = 4. 

\begin{figure}
    \centering
    \includegraphics[width=\linewidth]{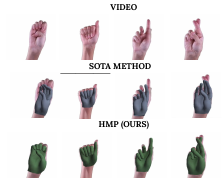}
    \caption{\textbf{3D hand pose and shape estimation on an in-the-wild video:} input video (top), \pymafx (middle), \methodname (bottom)} 
    \label{fig:in_the_wild_qual1}
\end{figure}

\begin{figure}
    \centering
    \includegraphics[width=\linewidth]{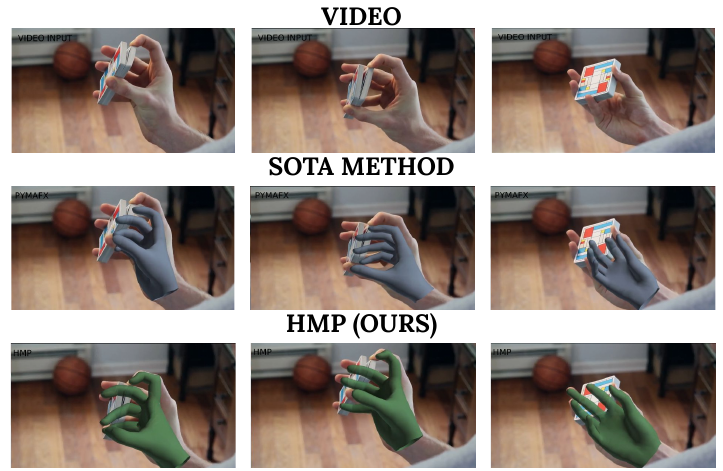}
    \includegraphics[width=\linewidth]{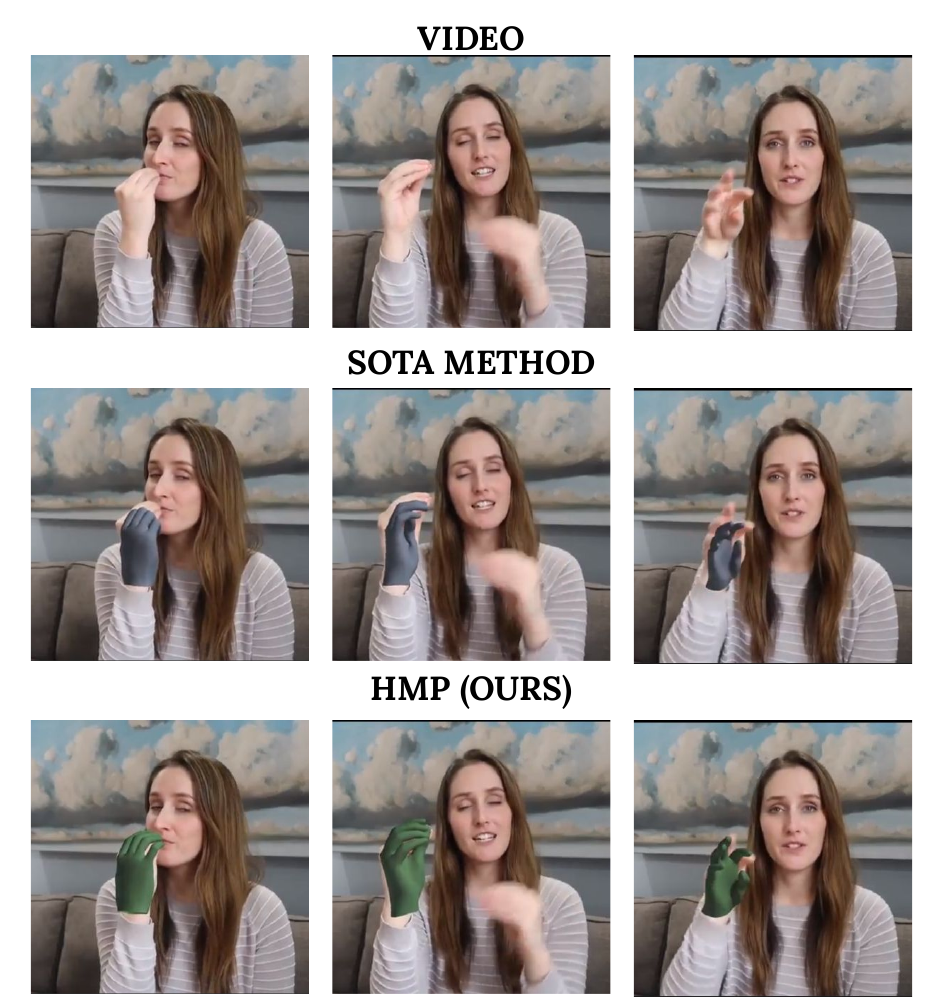}
    \caption{\textbf{3D hand pose and shape estimation on an in-the-wild video:} input video (top), \pymafx (middle), \methodname (bottom)} 
    \label{fig:in_the_wild_qual3}
\end{figure}

\begin{figure*}[t]
    \centering
    \includegraphics[width=\textwidth]{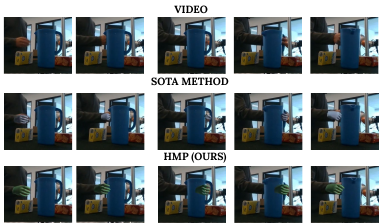}
    \caption{\textbf{3D hand pose and shape estimation on DexYCB videos:} input video (top), \pymafx (middle), \methodname (bottom)} 
    \label{fig:dexycb_qual}
\end{figure*}

\textbf{Qualitative Results:} We provide more qualitative results on DexYCB dataset and in-the-wild videos in Fig. \ref{fig:in_the_wild_qual1}, \ref{fig:in_the_wild_qual3}, and \ref{fig:dexycb_qual}. We refer the readers to our SupMat video for more results.

{}

{}


\section{Training A Hand Motion Prior}
\label{sec:hmp_training}

\textbf{Data Preprocessing:} We adopted a similar approach to data preprocessing as \cite{he2022nemf, rempe2021humor}. GRAB, TCDHands, and SAMP datasets in AMASS have hand articulation \cite{AMASS:2019}. We only train motion prior for the right hand. We first reflect left hand articulations in the dataset for data augmentation. Then all sequences are divided to motion clips of 128 timesteps. These clips are preprocessed to obtain processed data $\mathbf{X}$. For any timestep $t$, the processed data point $\mathbf{X}_t$ is, 
\begin{equation}
\mathbf{X}_t=\left(\begin{array}{llll}\mathbf{x}_t^p & \dot{\mathbf{x}}_t^p & \mathbf{x}_t^r & \dot{\mathbf{x}}_t^r\end{array}\right) \in \mathbb{R}^{J \times 15}.
\end{equation}
For a timestep $t$, $\mathbf{x}_t^r\in\mathbb{R}^{J\times3}$ denotes joint positions, $\dot{\mathbf{x}}_t^p\in\mathbb{R}^{J\times3}$ denotes joint velocities,
$\mathbf{x}_t^r\in\mathbb{R}^{J\times6}$ denotes hand pose in 6D rotation representation \cite{zhou2019continuity}, $\dot{\mathbf{x}}_t^r\in\mathbb{R}^{J\times3}$ denotes angular velocity. $J$ indicates the number of joints. 

\textbf{Architecture:} The architecture is adapted from \cite{he2022nemf}. We employed the Adam optimizer \cite{Adam}. The training process takes $\sim$10 hours on NVIDIA-A100 GPU. Our motion prior contains parameters contains $\sim$63.4 million parameters. Please refer to  Table \ref{tab:hmp_training} for hyperparameter values.

\begin{table}
    \centering
    \begin{tabular}{c|c}
       Hyperparameter & Value \\ 
       \toprule
       Epochs  & 1000 \\
       Batch Size & 16 \\
       Learning Rate  & 1e-4 \\
       Weight Decay & 1e-4 \\ 
       Latent Dimension & 1024 

    \end{tabular}
    \caption{Hyperparameter setting in motion prior training}
    \label{tab:hmp_training}
\end{table}
{}


\begin{figure*}[t]
    \centering
    \includegraphics[width=\textwidth]{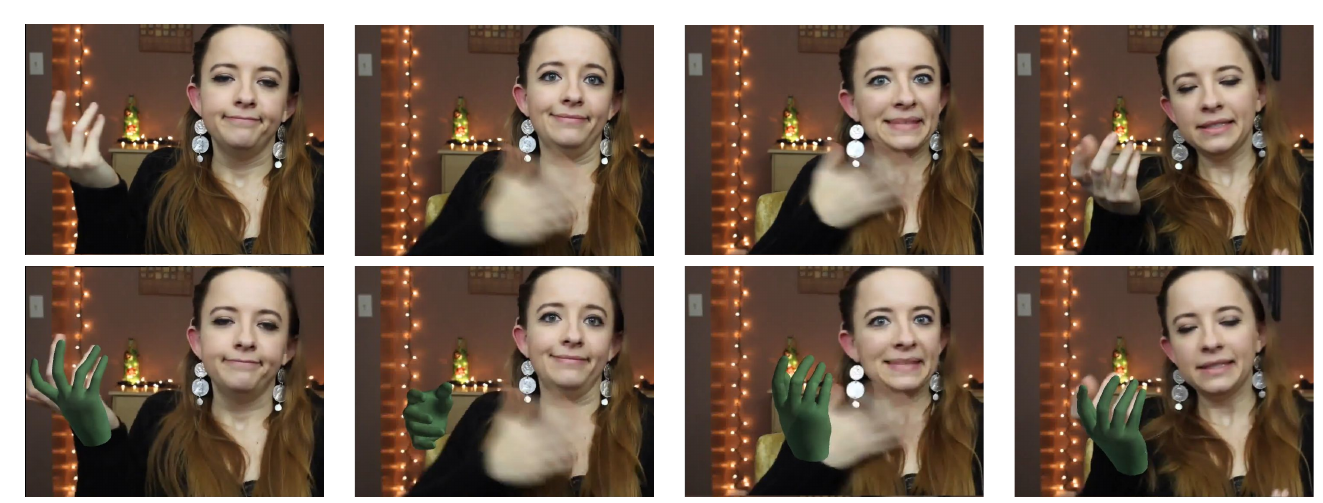}
    \caption{Failure in motion inbetweening due to the discontinuity in bounding box detection. Bounding boxes are detected only for the first and the final frame.} 
    \label{fig:bbox_discontinue}
\end{figure*}{}
\begin{figure*}[t]
    \centering
    \includegraphics[width=\textwidth]{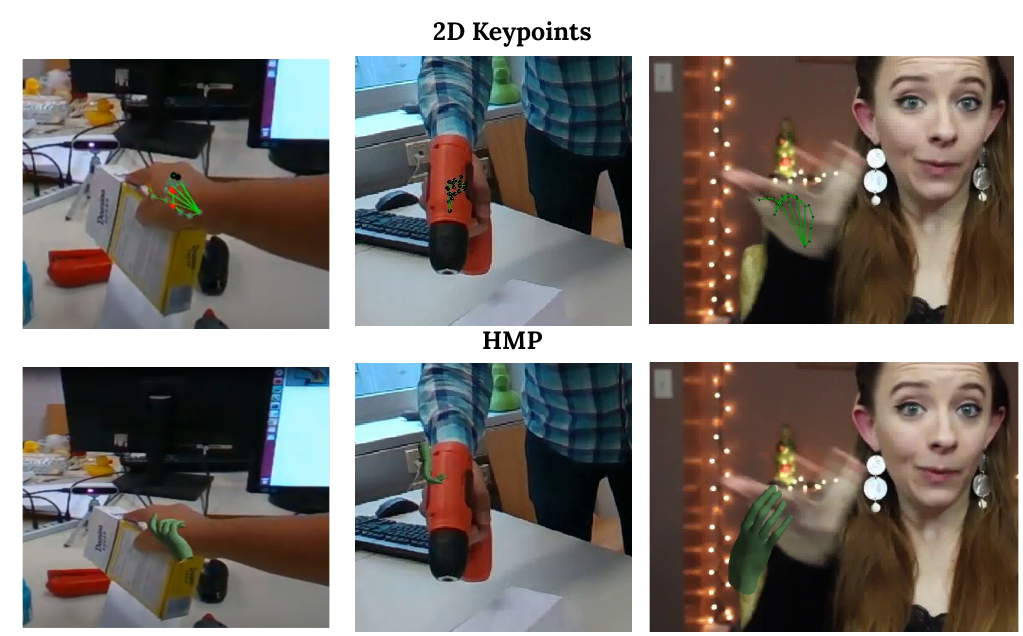}
    \caption{Failure cases caused by faulty keypoint detections. Keypoints detected (top), \methodname (bottom)} 
    \label{fig:keypoint_fail}
\end{figure*}{}
\section{Failure Cases}

\textbf{Keypoint Detection Failure:} One key cause for our method's failure is inaccurate keypoints. Under motion blur and occlusion, current state-of-the-art keypoint detectors tend to fail providing correct detections. Fig. \ref{fig:keypoint_fail} shows those cases along with our methods' output. 

\textbf{Bounding Box Discontinuity:} Another risk of failure originates from hand bounding box detection. Our method fails to interpolate and perform motion inbetweening. This usually happens with motion blur. An example can be seen in Fig. \ref{fig:bbox_discontinue}.



\clearpage
\balance 
{ \small
\bibliographystyle{ieee_fullname}

}